%% file: main.tex
\documentclass[11pt,a4paper]{article}
\usepackage{authblk} 
\usepackage[hyperref]{eacl2021}
\usepackage{times}
\usepackage{hyperref}
\usepackage{latexsym}
\usepackage{todonotes}
\usepackage{amsmath}

\usepackage{microtype}

\aclfinalcopy 
\def\aclpaperid{350} 

\usepackage{subfig}
\usepackage{booktabs}
\usepackage{pbox}
\usepackage{todonotes}
\usepackage{tabularx}
\newcolumntype{L}{>{\RaggedRight\arraybackslash}X}
\graphicspath{ {./images/} }

\author[1]{\bf Betty van Aken}
\author[1]{\bf Jens-Michalis Papaioannou}
\author[2]{\bf Manuel Mayrdorfer}
\author[2]{\\ \bf Klemens Budde}
\author[1]{\bf Felix A. Gers}
\author[1]{\bf Alexander Löser}
\affil[1]{Beuth University of Applied Sciences Berlin}
\affil[2]{Charité Berlin}
\affil[ ]{\tt \normalsize \{bvanaken,michalis.papaioannou,gers,aloeser\}@beuth-hochschule.de}
\affil[ ]{\tt \normalsize \{manuel.mayrdorfer,klemens.budde\}@charite.de}

\date{}

\begin{document}

\title{Clinical Outcome Prediction from Admission Notes\\using Self-Supervised Knowledge Integration}

\maketitle
\begin{abstract}
Outcome prediction from clinical text can \hbox{prevent} doctors from overlooking possible risks and help hospitals to plan capacities. We simulate patients at admission time, when decision support can be especially valuable, and contribute a novel \textit{admission to discharge} task with four common outcome prediction targets: Diagnoses at discharge, procedures performed, in-hospital mortality and length-of-stay prediction. The ideal system should infer outcomes based on symptoms, pre-conditions and risk factors of a patient. We evaluate the effectiveness of language models to handle this scenario and propose \textit{clinical outcome pre-training} to integrate knowledge about patient outcomes from multiple public sources. We further present a simple method to incorporate ICD code hierarchy into the models. We show that our approach improves performance on the outcome tasks against several baselines. A detailed analysis reveals further strengths of the model, including transferability, but also weaknesses such as  handling of vital values and inconsistencies in the underlying data.
\end{abstract}

\input{sections/Introduction.tex}
\input{sections/RelatedWork.tex}

\input{sections/ArchitectureFigure.tex}
\input{sections/Task.tex}
\input{sections/Method.tex}
\input{sections/Experiments.tex}
\input{sections/Discussion.tex}
\input{sections/Conclusion}

\section*{Acknowledgments}

We would like to thank Anjali Grover and Sebastian Herrmann for their support throughout the project.
Our work is funded by the German Federal Ministry for Economic Affairs and Energy (BMWi) under grant agreement 01MD19003B (PLASS) and 01MK2008MD (Servicemeister).

\bibliographystyle{acl_natbib}
\bibliography{citations}

\appendix
\input{sections/Appendix}

\end{document}

%% file: sections/Introduction.tex
\begin{figure*}[t!]
  \centering
  \includegraphics[width=160mm]{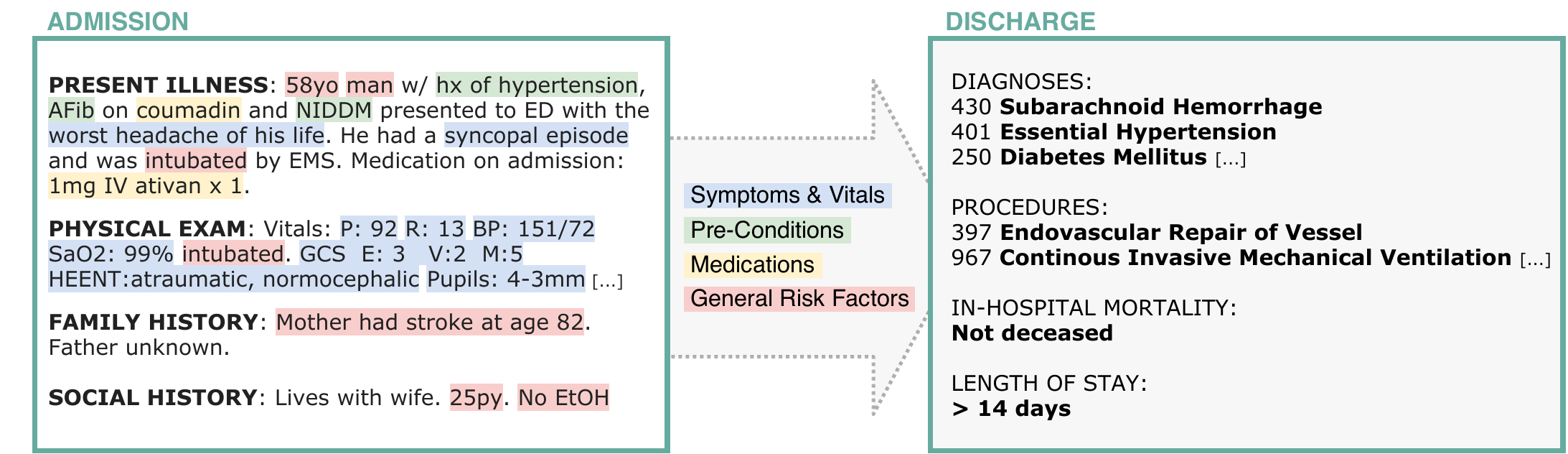}
  \caption{\textit{Admission to discharge} sample that demonstrates the outcome prediction task. The model has to extract patient variables and learn complex relations between them in order to predict the clinical outcome.}
\label{fig:note}
\end{figure*}

\section{Introduction}
\label{sec:intro}
Clinical professionals make decisions about patients under strong time constraints. The patient information at hand is often unstructured, e.g. in the form of clinical notes written by other medical personnel in limited time. Clinical decision support (CDS) systems can help in these scenarios by pointing towards related cases or certain risks. Clinical outcome prediction is a fundamental task of CDS systems, in which the patient's development is predicted based on data from their Electronic Health Record (EHR). In this work we focus on textual EHR data available at admission time.\\
Figure \ref{fig:note} shows a sample admission note with highlighted parts that -- according to medical doctors -- must be considered when evaluating a patient.

\paragraph{Encoding clinical notes with pre-trained \hbox{language} models.} Neural models need to extract relevant facts from such notes and learn complex relations between them in order to associate certain clinical outcomes. Pre-trained language models such as BERT \cite{bert} have shown to be able to both extract information from noisy text and to capture task-specific relations in an end-to-end fashion \cite{tenney,explainbert}. We thus base our work on these models and pose the following questions:

\begin{itemize}
    \item Can pre-trained language models learn to predict patient outcomes from their admission information only?
    \item How can we integrate knowledge about outcomes that doctors gain from medical literature and previous patients?
    \item How well would these models work in clinical practice? Are they able to interpret common risk factors? Where are they failing?
\end{itemize}

\paragraph{Simulating patients at admission time.} Existing work on text-based outcome prediction focuses on progress notes after a certain time of a patient's hospitalisation \cite{huang-clinical-bert}. This is mostly due to a lack of publicly available admission notes and poses some problems: 1) Doctors might miss specific outcome risks early in admission and 2) progress notes already contain information about clinical decisions made on admission time \cite{whats-in-a-note}. We propose to simulate newly arrived patients by extracting admission notes from MIMIC III discharge summaries. We are thus able to give doctors hints towards possible outcomes from the very beginning of an admission and can potentially prevent early mistakes. We can also help hospitals in planning resources by indicating how long a patient might stay hospitalised.

\paragraph{Integrating knowledge with specialised \hbox{outcome pre-training}.} \citet{pretraining} recently emphasized the importance of domain- and task-specific pre-training for deep neural models. Consequently we propose to enhance language models pre-trained on the medical domain with a task-specific \textit{clinical outcome pre-training}. Besides processing clinical language with idiosyncratic and specialized terms, our models are thus able to learn about patient trajectories and symptom-disease associations in a self-supervised manner. We derive this knowledge from two main sources: 1) Previously admitted patients and their outcomes. This knowledge is usually stored by hospitals in unlabelled clinical notes and 2) Scientific case reports and knowledge bases that describe diseases, their presentations in patients and prognoses. We introduce a method for incorporating these sources by creating a suitable pre-training objective from publicly available data.

\paragraph{Contributions.} We summarize the major contributions of this work as follows:\\
1) A novel task setup for clinical outcome prediction that simulates the patient's admission state and predicts the outcome of the current admission.\\
2) We introduce self-supervised \textit{clinical outcome pre-training}, which integrates knowledge about patient outcomes into existing language models.\\
3) We further propose a simple method that injects hierarchical signals into ICD code prediction.\\
4) We compare our approaches against multiple baselines and show that they improve performance on four relevant outcome prediction tasks with up to 1,266 classes. We show that the models are transferable by applying them to a second public dataset without additional fine-tuning.\\
5) We present a detailed analysis of our model that includes a manual evaluation of samples conducted by medical professionals.

%% file: sections/RelatedWork.tex
\section{Related Work}

\paragraph{Using clinical notes for outcome prediction.} \citet{whats-in-a-note} studied the predictive value of clinical notes with simple approaches such as bag-of-words. Recent work increasingly applies neural models to compensate for the noisy nature of the data and the complexity of patterns. \citet{hashir-mortality} used both convolutional and
recurrent layers for outcome prediction, while \citet{attention-notes} and \citet{MNN} proposed attention-based approaches. \citet{pretraining-substance-misuse} explored pre-training as a strategy to mitigate data sparsity in clinical setups. \citet{multitask-mortality} and \citet{multitask-patients} further showed that outcome prediction benefits from a multitask setup. In contrast to earlier work we apply neural models to admission notes in an \textit{admission to discharge} setup.

\paragraph{Pre-trained language models for the clinical \hbox{domain.}}
While pre-trained language models are successful in many areas of NLP, there has been little work on applying them to the clinical domain \cite{pretrained-models-survey}. \citet{alsentzer-clinical-bert} and \citet{huang-clinical-bert} both pre-trained BERT-based models on clinical data. They evaluated their work on readmission prediction and other NLP tasks. We are the first to evaluate pre-trained language models on multiple clinical outcome tasks with large label sets. We further propose a novel pre-training objective specifically for the clinical domain.

\paragraph{Prediction of diagnoses and procedures.} The majority of work on diagnosis and procedure prediction covers either single diagnoses \cite{deep-ehr,mime} or coarse-grained groups \cite{self-attention-enhance,sushil}. We argue that models should predict diseases and procedures in a fine-grained manner to be beneficial for doctors. Thus we use all diagnosis and procedure codes from the data for our outcome prediction tasks.

\paragraph{ICD coding vs. outcome prediction.}
There is a variety of work in the related field of automated ICD coding \cite{icd-coding1,icd-coding2}. \citet{bert-xml} recently presented a model able to identify up to 2,292 ICD codes from text. However, ICD coding differs from outcome prediction in the way that diseases are directly extracted from text rather than inferred from symptom descriptions and patient history. We further discuss this distinction in Section \ref{section:discussion}.

%% file: sections/ArchitectureFigure.tex
\begin{figure*}[t!]
  \centering
  \includegraphics[width=160mm]{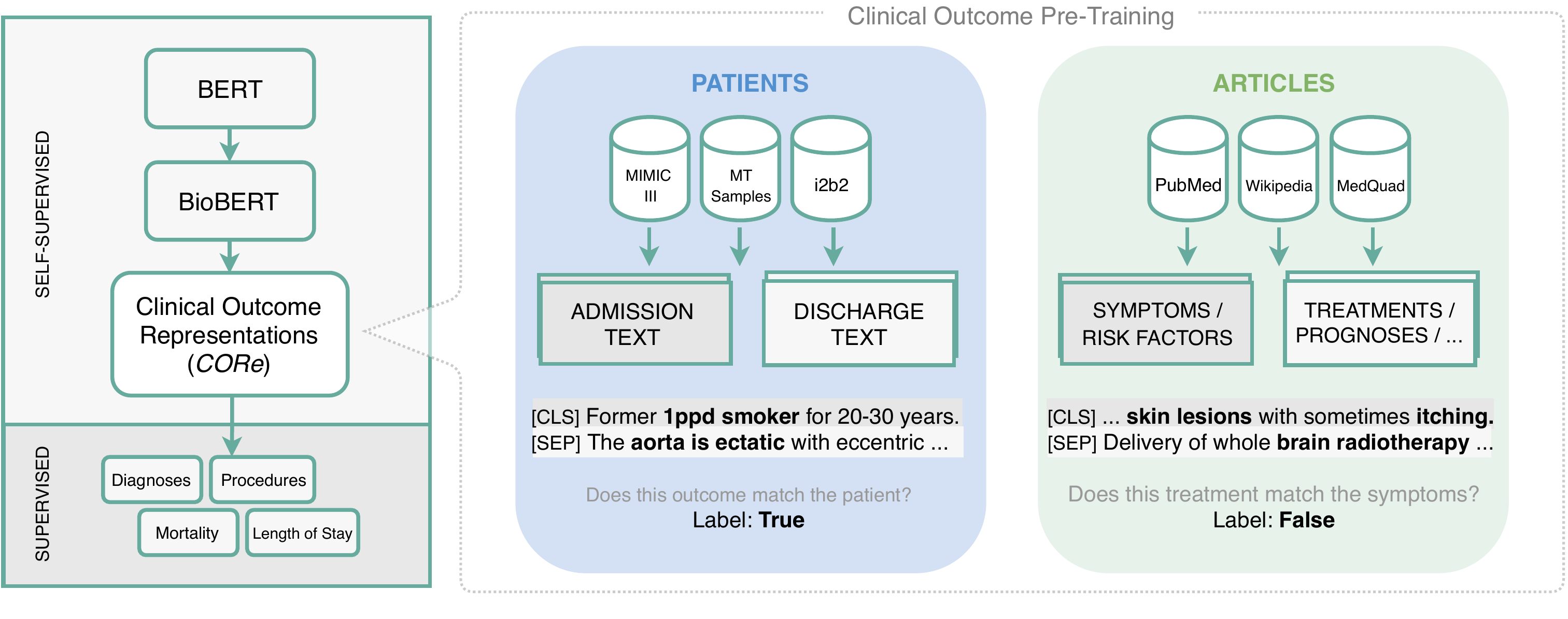}
  \caption{Schematic demonstration of \textit{clinical outcome pre-training}. Sources of clinical knowledge are complete patient notes and medical articles. Based on that we create a self-supervised learning objective that teaches relations between symptoms, risk factors and outcomes.}
\label{fig:architecture}
\end{figure*}

%% file: sections/Task.tex
\begin{table}[b]
    \resizebox{\columnwidth}{!}{
      \begin{tabular}{ |c | c| c| c| }
              \hline
                 \multicolumn{4}{|c|}{\textbf{Admission Notes Statistics}}\\ \hline
                  avg & std & avg & std\\ 
                  (words / doc) & (words / doc) & (sent / doc) & (sent / doc)\\ \hline
                  396.3 & 233.3 & 32.5 & 23.1\\
            \hline
      \end{tabular}
      }
      \caption{Numbers of words / sentences in MIMIC III admission notes. We see a high variation in length.}
      \label{mimic_stats}
\end{table}

\begin{table}[t]
    \resizebox{\columnwidth}{!}{
      \begin{tabular}{ |c | c | c | c | c | c | c | c | }
              \hline
                 \multicolumn{8}{|c|}{\textbf{Multi-label tasks:} ICD-9 codes per dataset split}\\ \hline
                 \multicolumn{4}{|c|}{Diagnoses} & \multicolumn{4}{c|}{Procedures}\\ \hline
                 \textbf{Total} & Train & Val & Test & \textbf{Total} & Train & Val & Test \\ \hline
                 \textbf{1,266} & 1,201 & 906 & 1,031 & \textbf{711} & 672 & 476 & 563 \\
            \hline
      \end{tabular}
    }
      \caption{Distribution of ICD-9 codes per dataset split (patient-wise). Note that very rare codes do not appear in each split of the dataset.}
      \label{multilabel_task_stats}
\end{table}

\begin{table}[t]
    \resizebox{\columnwidth}{!}{%
      \begin{tabular}{ |c | c | c | c | c | c | }
              \hline
                 \multicolumn{6}{|c|}{\textbf{Single-label tasks:} Samples per class}\\ \hline
                 \multicolumn{2}{|c|}{Mortality} & \multicolumn{4}{c|}{Length of Stay (in days)}\\ \hline
                 0 & 1 & $\leq$ 3 & $>$ 3 \& $\leq$ 7 & $>$ 7 \& $\leq$ 14 & $>$ 14 \\ \hline
                 43,609 & 5,136 & 5,596 & 16,134 & 13,391 & 8,488 \\
            \hline
      \end{tabular}
    }
      \caption{Distribution of labels for \textit{Mortality Prediction} and \textit{Length of Stay} task. Both tasks have unbalanced class distributions.}
      \label{singlelabel_task_stats}
\end{table}

\section{Clinical \emph{Admission to Discharge} Task}
Clinical outcome prediction can be defined in different ways. We approach the task from a doctor's perspective and predict the outcome of a current admission from the time of the patient's arrival to the hospital unit. We describe our setup as follows.

\subsection{Clinical Notes from MIMIC III}
As our primary data source, we use the freely-available MIMIC III v1.4 database \cite{mimic}. It contains de-identified EHR data including clinical notes in English from the Intensive Care Unit (ICU) of Beth Israel Deaconess Medical Center in Massachusetts between 2001 and 2012. We focus our work on discharge summaries in particular and the outcome information associated with an admission. Similar to previous work, we filter out notes about newborns and remove duplicates.

\subsection{Creating Admission Notes from Discharge Summaries} \label{section:admission_time}
The state of a patient is commonly summarized in an ongoing document, which finally concludes in a discharge summary. Since we want to support clinical decisions from the beginning of a patient's stay, we simulate the state of the patient's document at admission time. We thus filter the document by sections that are known at admission such as: \textit{Chief complaint, (History of) Present illness, Medical history, Admission Medications, Allergies, Physical exam, Family history} and \textit{Social history}. We further describe the filtering in Appendix \ref{app:filter}. Our approach results in 48,745 admission notes. As shown in Table \ref{mimic_stats} the notes contain about 400 words on average.
The selection of admission sections as well as the resulting structure of the notes were verified by medical doctors.\\
This newly created admission dataset enables us to make predictions on the outcome of a current admission. At inference time, doctors can then use the model's predictions on textual data from newly arrived patients.

\subsection{Outcome Prediction Tasks} \label{section:outcome_tasks}
We select four relevant tasks for outcome prediction in consultation with medical professionals. All tasks take admission notes as input.

\paragraph{Diagnosis prediction.} A main goal of clinical outcome prediction is to support medical professionals in the process of differential diagnosis. We thus take all diagnoses associated with an admission into account and frame the task as an extreme multi-label classification. Diagnoses are encoded as ICD-9 codes in the MIMIC III database. Following \citet{medgan}, we group ICD-9 diagnosis codes from the database from 4- into 3-digit codes to reduce complexity while still obtaining granular suggestions. This results in a total of 1,266 diagnosis codes, which are distributed over our dataset splits as shown in Table \ref{multilabel_task_stats}. The labels are power-law distributed with a long tail of very rare codes.

\paragraph{Procedure prediction.} Procedures are either diagnostics or treatments applied to a patient during a stay. Similarly to diagnosis prediction, this is an extreme multi-label task. We again group the ICD-9 codes from the MIMIC III database into 3-digit codes. In total there are 711 procedure codes labelled in the database in a power law distribution similar to the diagnosis codes.

\paragraph{In-hospital mortality prediction.} Predicting a patient's mortality risk is a fundamental part of the triage process. In-hospital mortality in particular describes whether a patient died during the current admission and is a binary classification task. The percentage of deceased patients in the data is around 10\% (see Table \ref{singlelabel_task_stats}). As some notes contain direct indications of mortality such as \textit{patient deceased} within the admission sections, we apply an additional filter for those terms.

\paragraph{Length-of-stay prediction.} The duration of an ICU stay is an important information for hospitals in order to plan allocations of resources. We group patients into four major categories regarding their length of stay: \textit{Under 3 days, 3 to 7 days, 1 week to 2 weeks, more than 2 weeks.} These categories were recommended by medical doctors in order to make the results as useful as possible in clinical practice. Table \ref{singlelabel_task_stats} shows the samples per class.

%% file: sections/Method.tex
\section{Integrating Clinical Knowledge Into Language Models}
We propose \textit{clinical outcome pre-training}, a way to integrate knowledge about clinical outcomes into pre-trained language models. We further introduce an additional step to incorporate ICD code hierarchy into our multi-label classification tasks.\footnote{The code to recreate the experiments and datasets described in this paper is accessible at: \url{https://github.com/bvanaken/clinical-outcome-prediction}}

\subsection{Clinical Outcome Pre-Training}
\paragraph{Motivation.} 
Language model pre-training has shown to be of use in specialised domains like the clinical \cite{alsentzer-clinical-bert, huang-clinical-bert}. However, these models lack knowledge about patient trajectories and symptom-diagnosis relations, because their training is  focused on learning language characteristics.\\
We develop an additional pre-training step that produces \textit{Clinical Outcome \hbox{Representations} (CORe)} in order to teach the model relations between symptoms, risk factors and clinical outcomes.
Much of this knowledge is present and publicly available, e.g.~in knowledge bases like Wikipedia or publication archives like PubMed. Another source is available to hospitals in the form of unlabelled clinical notes from previous patients. The suggested outcome pre-training is a way to use this knowledge to improve the model's capabilities in predicting clinical outcomes as described in \ref{section:outcome_tasks}.\\
Corresponding to the way doctors gain their knowledge from both experience and medical literature, we incorporate knowledge from complete patient notes (including discharge information) and medical articles.

\paragraph{Training objective.} Our proposed training objective (Figure \ref{fig:architecture}) is strongly related to the Next Sentence Prediction (NSP) task introduced by \citet{bert}. In NSP the model gets two sentences as an input and predicts whether the second follows the first sentence. This way models such as BERT learn relations between sentences. We convert this setting so that the model instead learns relations between admissions and outcomes.\\
From common sections in patient notes, we create two categories: Sections that are created at admission $A$ and sections that are created after admission, e.g.~at discharge time $D$.
Given a patient note $N$, we split it into sections $A_N \in A$ and $D_N \in D$. We remove all other sections. We then sample token sequences from these sections to get $t_{N, 1\dots k} \in A_N$ and $t'_{N, 1\dots k} \in D_N$, where $k$ is randomly set between 30 and 50 tokens. We then train the model to maximize $P(Same\_Patient|X_{N\_N})$ and $P(Other\_Patient|X_{N\_M})$ with
\begin{equation}
\begin{alignedat}{2}
    X_{N\_N} = Enc(t_{N, 1\dots k}, t'_{\textbf{N}, 1\dots k})
    \\
    X_{N\_M} = Enc(t_{N, 1\dots k}, t'_{\textbf{M}, 1\dots k})
\end{alignedat}
\end{equation}
with $M$ being a randomly sampled document from the same batch and $Enc$ referring to the BioBERT encoding. As in the original NSP setting, we apply negative sampling ($X_{N\_M}$)  for 50\% of examples. We apply the same strategy on medical articles and case reports, so that $A$ represents sections describing symptoms and risk factors, and $D$ represents sections that describe outcomes of a disease or case.

\paragraph{Data sources.} We create the pre-training dataset from multiple public sources. To integrate knowledge that doctors gain from previous patients and medical literature, we create two groups of sources: \\1) \textit{Patients}, which includes 32,721 discharge summaries from the MIMIC III training set, 5,000 publicly available medical transcriptions from the MTSamples website \footnote{https://mtsamples.com} and 4,777 clinical notes from the i2b2 challenges 2006-2012\footnote{We exclude notes from the 2014 De-identification and Heart Disease Risk Factors Challenge in order to use this set for evaluation as described in Section \ref{sec:i2b2}.} \cite{i2b2-2006,i2b2-2006-2,i2b2-2009,i2b2-2009-2,i2b2-2010,i2b2-2011,i2b2-2012,i2b2-2012-2}.\\
2) \textit{Articles}, composed of 9,335 case reports from PubMed Central (PMC), 2,632 articles from Wikipedia describing diseases and 1,467 article sections from the MedQuAd dataset \cite{medquad} extracted from NIH websites such as cancer.gov.\\
While \textit{Patients} samples contain unaudited practical knowledge, \textit{Articles} samples are built from verified general medical knowledge such as peer-reviewed studies. The sources are therefore substantially different and we evaluate their individual effect on performance in Section \ref{section:results}.

\paragraph{Data preparation.} 
We create admission ($A_N$) and discharge parts ($D_N$) of the documents based on section headings. 
We define common sections belonging to the admission part and those belonging to the discharge part similar to the method described in Section \ref{section:admission_time}. We ignore sections that cannot be categorized. For section heading extraction from MIMIC III discharge summaries and MTSamples transcriptions, we apply simple rule-based approaches, which is feasible because the notes are well-structured. For Wikipedia we use headings from the WikiSection dataset \cite{sector} filtered for disease articles only. For PubMed Central we similarly use the PubMedSection dataset \cite{language-modeling-enough} and filter for section headings that indicate case reports. As i2b2 notes are less well-structured in comparison to MIMIC III discharge summaries, we use a classifier as proposed by \citet{i2b2-section-prediction} to determine which section a sentence belongs to. The classifier is trained on an annotated set of i2b2 notes and then applied to all other notes.

\subsection{ICD+: Incorporation of ICD Hierarchy} 
\begin{figure}[t!]
  \includegraphics[width=0.48\textwidth]{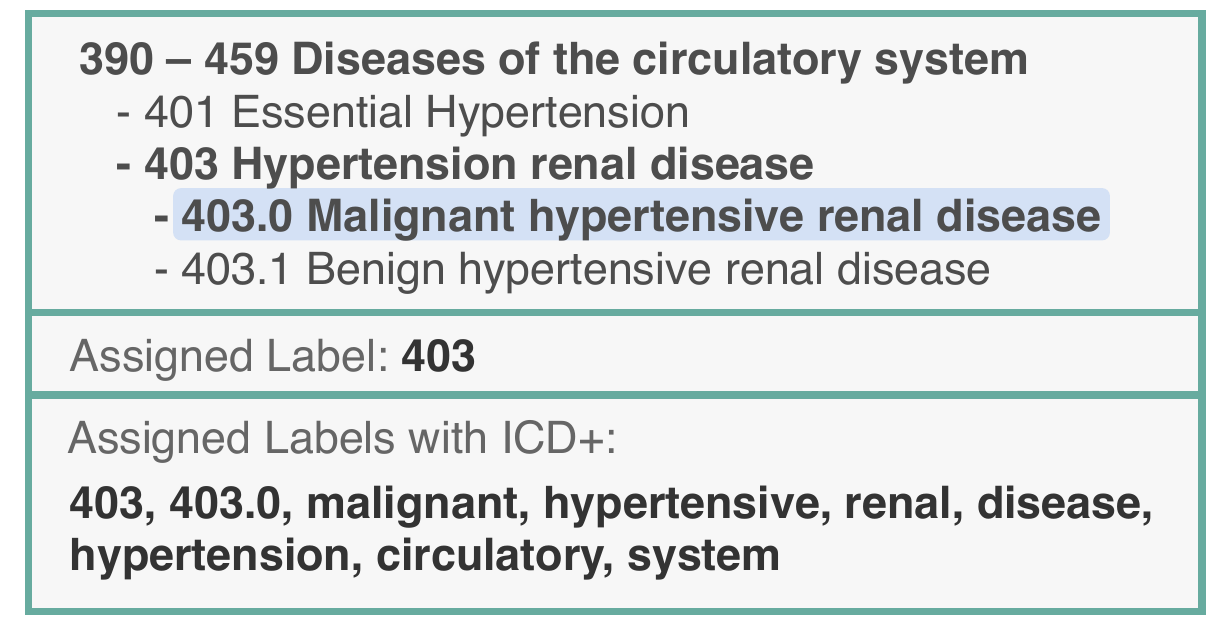}
  \caption{Example of \textit{ICD+} labelling. \textit{Malignant hypertensive renal disease} is assigned to nine codes \hbox{(bottom row)} that inform about the type and group of the disease.}
\label{fig:icd_plus}
\end{figure}

\paragraph{Medical knowledge in ICD labels.} Diagnosis and procedure prediction requires the model to predict ICD-9 codes in a multi-label manner. ICD-9 codes are hierarchically ordered into associated groups. Figure \ref{fig:icd_plus} shows the  code hierarchy for \textit{Malignant hypertensive renal disease} with the ICD-9 code \textit{403.0}. The diagnosis has two parent groups namely \textit{Hypertension renal disease} and \textit{Diseases of the circulatory system}.
Diagnoses or procedures in the same group often share similar medical characteristics, therefore hierarchical relations of a labelled code can be valuable information. This medical information is currently not integrated into the model. The same holds for words describing the ICD-9 codes, that often represent further important signals, such as the words \textit{renal} or \textit{malignant}.
\paragraph{Enhancing training with useful additional \hbox{signals}.}
We propose a simple method, \textit{ICD+}, to incorporate both associated groups and words into the model weights: Instead of only classifying 3-digit codes (as mentioned in \ref{section:outcome_tasks}), we let the model additionally predict the 4-digit codes and the bag of associated words with a code and its parent groups. In order to create the bag of words per code, we use the descriptions of ICD-9 codes from MIMIC III and remove all stop words. As shown in Figure \ref{fig:icd_plus}, the \textit{ICD+} method assigns eight additional labels to the example diagnosis and therefore supplies the model with  further information about the diagnosis during training.\\By increasing the amount of labels per sample, we integrate relevant medical knowledge and enable the model to learn implicit relations between codes and code groups that share certain words. We evaluate the effectiveness of \textit{ICD+} in Section \ref{section:experiments}.

\begin{table*}[t]
    \resizebox{2.08\columnwidth}{!}{%
\begin{tabular}{@{}lllll@{}}
\hline
 & Diagnoses & Procedures & In-Hospital Mortality &  Length-of-Stay \\
 & (1266 classes) & (711 classes) & (2 classes) & (4 classes) \\
 \hline
BOW \cite{whats-in-a-note} & 75.87 & 77.47 & 79.15 & 65.83 \\
Embeddings \cite{whats-in-a-note} & 75.16 & 76.72 & 79.94 & 66.78 \\
CNN \cite{multitask-mortality} & 61.18 &  73.13 & 75.50 & 64.49 \\ \hline
BERT Base \cite{bert} & 82.08 & 85.84 & 81.13 & 70.40 \\
ClinicalBERT \cite{huang-clinical-bert} & 81.99 &  86.15 & 82.20 & 71.14 \\ 
\textit{DischargeBERT} \cite{alsentzer-clinical-bert} & \textit{82.86} & \textit{87.09} & \textit{\textbf{84.51}} & \textit{71.73}  \\ 
BioBERT Base \cite{biobert} & 82.81 &  86.36 & 82.55 & 71.59 \\ \hline
BioBERT ICD+ & 83.17 & 87.45 & - & -  \\
CORe Articles \small{(w/o ICD+)} & 83.46 \small{(82.89)} & 87.43 \small{(86.75)} & 83.64 & 71.99 \\
CORe Patients \small{(w/o ICD+)} & 83.41 \small{(83.40)} & \textbf{88.37} \small{(86.60)} & 83.60 & 71.96 \\
CORe All \small{(w/o ICD+)} & \textbf{83.54} \small{(83.39)} & 87.65 \small{(87.15)} & \textbf{84.04} & \textbf{72.53} \\ \hline

\end{tabular}
}
\caption{Results on outcome prediction tasks in macro-averaged \% AUROC. The \textit{CORe} models outperform the baselines, \textit{ICD+} adds further improvement (values in parentheses are ablation results without \textit{ICD+}). DischargeBERT results are printed in italic because the model has seen all test data during pre-training and is therefore slightly advantaged.}
\label{table:results}
\end{table*}

%% file: sections/Experiments.tex
\section{Experimental Evaluation}
\label{section:experiments}

\subsection{Training Clinical Outcome Representations} We pre-train the \textit{CORe} model on top of BioBERT weights\footnote{We choose BioBERT as the base for our model because it outperforms BERT on medical tasks and has not seen data from our test set during pre-training unlike DischargeBERT.}. We then fine-tune the model separately on the four outcome tasks. We use the same training regimen for both pre-training and fine-tuning: We tokenize the texts with WordPiece tokenization and truncate them to 512 tokens, due to the limited context length of the pre-trained models. We use early stopping and tune hyperparameters as described in Appendix \ref{app:hyperparameter}.

\subsection{Baseline Models} \label{section:baselines} In the following, we introduce the baseline models that we evaluate on the novel outcome prediction tasks. In order to understand the abilities of pre-trained language models we compare their performance against more traditional approaches. The first three models (\textit{BOW, word embeddings, CNN}) are trained using the hyperparameters proposed by the authors for outcome prediction tasks. The language models are fine-tuned the same way as the \textit{CORe} model. 

\paragraph{Bag-of-Words.}
\citet{whats-in-a-note} shows that a simple bag-of-words (BOW) approach can outperform more complex models on tasks like mortality prediction. We thus include their approach in our evaluation. We adopt their training setting except that we consider 200 instead of 20 top tf-idf words in order to make the model converge.

\paragraph{Pre-trained word embeddings.} \citet{whats-in-a-note} further propose the use of pre-computed word embeddings that were trained on MIMIC III data. We use the same setting as for the BOW approach and fit a support vector machine classifier on the clinical outcome tasks.

\paragraph{Convolutional Neural Network (CNN).}
\citet{multitask-mortality} built a neural network for mortality prediction with two hierarchical convolutional layers at the word and sentence levels and then aggregated it to a patient level representation. We follow their approach to evaluate the model on our four \textit{admission to discharge} tasks.

\paragraph{BioBERT.}
Following the success of BERT, \citet{biobert} further pre-trained the model on biomedical research articles from PubMed using abstracts and full-text articles. They reported improved performance on a range of biomedical text mining tasks.

\paragraph{ClinicalBERT and DischargeBERT.} We further evaluate two public language models pre-trained on the clinical domain, with MIMIC III data in particular. \citet{huang-clinical-bert} pre-trained a BERT Base model on 100,000 random clinical notes (ClinicalBERT) while \citet{alsentzer-clinical-bert} further pre-trained BioBERT on all discharge summaries from MIMIC III (we refer to the model as DischargeBERT for simplicity).

\subsection{Results on MIMIC III Admission Notes}
\label{section:results}
Table \ref{table:results} shows performances in (macro-averaged) area under the receiver operating characteristic curve (AUROC). We report scores of the \textit{CORe} model trained only on \textit{Articles}, \textit{Patients} and in a combined training setting \textit{CORe All}. We evaluate diagnosis and procedure prediction both with and without the \textit{ICD+} method on BioBERT and the \textit{CORe} models. In both scenarios we evaluate on 3-digit ICD codes only, in order to maintain comparability between the methods.

\paragraph{Pre-trained models outperform baselines.}
We see that the evaluated pre-trained language models clearly outperform the \textit{BOW}, \textit{word embeddings} and \textit{CNN} approaches. We further observe that the \textit{CORe} models improve scores on all tasks in comparison to the baseline models, except for DischargeBERT that reaches a higher score in mortality prediction -- probably affected by its exposure to the test data. This shows that even though the language models are trained on similar data (e.g.~PubMed and/or clinical notes), the specific \textit{outcome pre-training} improves the model's ability to predict clinical outcome targets. Pre-training on \textit{Patients} and \textit{Articles} achieve similar improvements over the baselines, while the combined training is the most effective. An exception is the procedure prediction, where pre-training on \textit{Patients} achieves the highest score. A probable reason is that procedures are documented in more detail in clinical notes, especially since our selection of medical articles focuses on diseases rather than procedures.

\paragraph{Predicting mortality risk is easier than length of stay.} We see that the models reach higher scores in the binary mortality task than in length of stay prediction. Even a simple \textit{BOW} approach can reach a relatively high score, which indicates that most of the notes contain clear hints towards an increased mortality risk. On the other hand, the length of stay task is difficult due to the many factors that can contribute to the length of a patient's stay after the admission, including nonclinical factors such as the patient's insurance situation \cite{los-factors}.

\paragraph{ICD hierarchy improves diagnosis and procedure predictions.} Table \ref{table:results} shows an ablation test without the \textit{ICD+} method (in parentheses). We see that both the BioBERT model and the \textit{CORe} models  improve when incorporating code hierarchy and relations through \textit{ICD+} into the training process. This is especially visible for ICD procedures, where the hierarchical and textual information, e.g. that a \textit{Nephropexy} is an \textit{operation} on the \textit{kidney} can add important signals during training. 

\begin{table}[b!]
        \begin{tabularx}{\columnwidth}{Xr}
            \hline
             & i2b2 Diagnoses\\ \hline
            BioBERT ICD+ & 80.43  \\
            CORe Articles & 81.46 \\ 
            CORe Patients  & \textbf{82.31} \\
            CORe All  & 81.15 \\ \hline
        \end{tabularx}
    \caption{Results on i2b2 diagnosis prediction task (5 classes) in \% AUROC. The models reach similar results as on the MIMIC III data, indicating their transferability to other data sources without additional fine-tuning.}
    \label{table:i2b2_results}
\end{table}

\subsection{Model Transferability: Cross-Verification on i2b2 Clinical Notes}
In order to verify that the fine-tuned models are transferable to ICU data from other sources, we apply it to data from the i2b2 De-identification and Heart Disease Risk Factors Challenge \cite{i2b2-2014}. We convert the clinical notes to admission notes as further described in Appendix \ref{app:i2b2-prep}, which results in 1,118 samples labelled with up to five ICD-9 codes.

\paragraph{Models generalize to i2b2 data.} 
\label{sec:i2b2}
We apply our MIMIC III-based models to predict diagnosis codes for the i2b2 notes without further fine-tuning. We then evaluate based on whether the predictions contain the five mentioned ICD-9 codes. The results in macro-averaged \% AUROC are shown in Table \ref{table:i2b2_results}. Even though the clinical notes differ from the MIMIC III notes in structure and writing style, the tested models are mostly able to identify the conditions. The scores are comparable to the MIMIC III results, which shows that the models are able to generalise on data from different sources such as other hospitals.

%% file: sections/Discussion.tex
\section{Discussion and Findings}
\label{section:discussion}
Clinical outcome prediction is a sensitive task. We therefore conduct an extensive analysis on the \textit{CORe All} model including a manual error analysis by medical doctors on 20 randomly chosen samples to understand how the model would perform in clinical practice.\footnote{Our demo application used for this analysis is available at: \url{https://outcome-prediction.demo.datexis.com}}

\begin{figure}
\centering
  \includegraphics[width=0.5\textwidth]{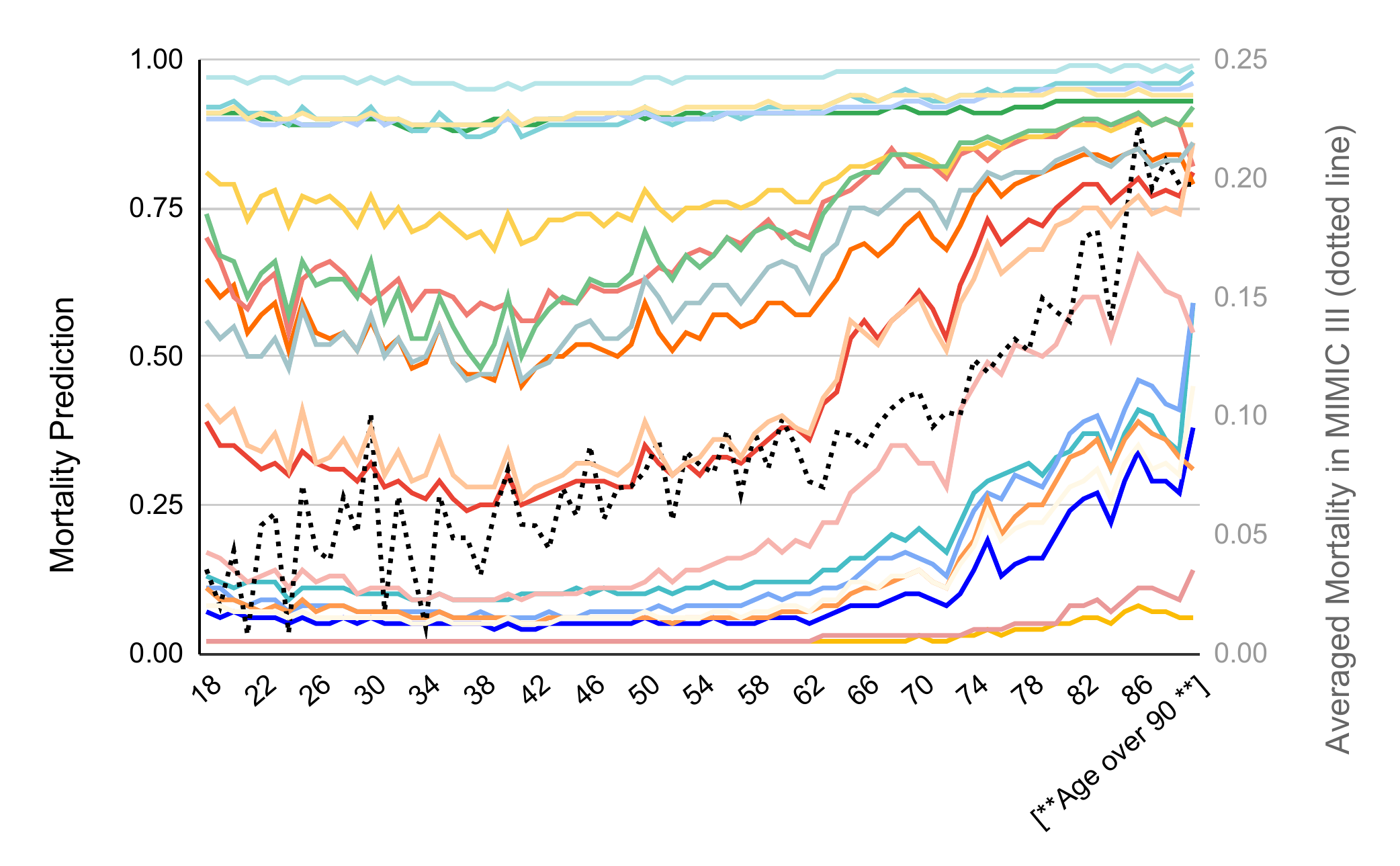}
  \caption{Impact of age on mortality prediction on 20 random samples. Mortality risk and age mostly increase proportionally as intended, with certain peaks that might indicate unintended biases in the data.}
\label{fig:age}
\end{figure}

\begin{table}[t!]
\begin{tabular}{@{}lr@{}}
\hline
 & \% AUROC\\ \hline
All Diagnoses & 83.54 \\
Diagnoses \textbf{Mentioned} in Text & \textbf{87.10} \\
Diagnoses \textbf{Not Mentioned} in Text & 82.35 \\\hline

\end{tabular}
\caption{Analysis of the impact of directly mentioned diagnoses on the diagnosis prediction task. Mentioned diagnoses are detected more reliably. Though on unmentioned diagnoses, scores only see a small decrease compared to the overall score.}
\label{table:extract_results}
\end{table}

\subsection{A Closer Look at the Model's Abilities}

\paragraph{Does the model mainly extract already present diagnoses?} We observe that a majority of coded diseases are already mentioned in the admission text. This is mainly due to chronic diseases (e.g. \textit{diabetes mellitus}) or to conditions that were identified prior to the ICU admission (e.g. in the emergency ward). We want to know if our model is also able to predict diagnoses that are not mentioned in the text. We annotate the admission texts with ICD-9 diagnosis codes with the methodology described by \citet{silver_mimic}. We then evaluate on codes that were explicitly mentioned in the text and those that were not. Table \ref{table:extract_results} shows that the model indeed extracts many diagnoses directly from the text and thus reaches a higher score on mentioned diagnoses. On the other hand, we see that the performance on non-mentioned diagnoses does drop only slightly, indicating that the model has also learned to predict non-mentioned diagnoses.

\paragraph{How does age and gender impact predictions?} Age and gender are common risk factors with significant impact on the potential clinical outcome of a patient. We want our models to learn that impact without overestimating it. We test the model's behaviour by switching age and gender throughout 20 random samples and analyse how the mortality prediction changes. For each sample we manually switch the age mention and iterate over it from 18 until [**Age over 90**]\footnote{De-identified age information for patients older than 89.}. Figure \ref{fig:age} shows that the analysed samples show a high variation in mortality risk and that age only impacts the prediction partially. In all cases the prediction increases with age -- as expected from a medical perspective. We also observe some peaks without a medical reason that are caused by the mortality of certain age groups in the original data (black dotted line). This demonstrates how the model does not follow medical reasoning but merely statistic observations. We similarly switch the gender mention and all pronouns in the texts and observe that mortality prediction for male patients is increased by 5\% on average, consistent with medical rationale.

\paragraph{Where is the model failing?}  
\begin{enumerate}
    \item \textbf{Negation}: While our error analysis depicts that negation does not generally falsify the model's predictions, we find single samples in which especially medical-specific negations, such as \textit{\hbox{abstinent} from alcohol}, are misinterpreted by the model, e.g. into \textit{alcohol \hbox{dependence} syndrome}.
    \item \textbf{Numerical data}: \citet{bert-numbers} show BERT's inabilities to interpret numbers. We observe this in the case that the model does not interpret life-threatening vital values (such as temperature over 105$^{\circ}$F) as an increased mortality risk. Clinical notes contain many such relevant values, thus improving the encoding of such data is an important goal for future work.
\end{enumerate}

\subsection{There is no Ground Truth in Clinical Data}

\paragraph{Incomplete and inconsistent labels.} Our error analysis reveals that 60\% of the analysed samples are partially under-coded. They contain indicators for a diagnosis or procedure but miss the corresponding ICD-9 code. This is consistent with results from \citet{silver_mimic} showing that MIMIC III is up to 35\% under-coded. Additionally we find that procedures that are almost always performed in the ICU such as \textit{Puncture of vessel} are often coded inconsistently. While a doctor can infer these labels with medical common sense, they pose a challenge to our models. We therefore suggest a critical view towards the data and welcome additional clinical datasets to compensate for noisy labels.

\paragraph{Multiple possible outcomes.} 85\% of analysed samples contain false positive predictions that the doctors still consider medically reasonable. This demonstrates that there are many possible clinical pathways and that some might not be foreseeable at admission time. We also see many cases in which the information in the clinical note is not sufficient and therefore allows multiple interpretations. For future work, we propose including further EHR data as suggested by \citet{notes-timeseries} to extend the patient representation in these scenarios.

%% file: sections/Conclusion.tex
\section{Conclusion}
We reframe the task of clinical outcome prediction to consider the admission state of a patient and support doctors in their initial decision process. We show that current state-of-the-art language models outperform selected baselines on this task and present methods for further improvement: \textit{Outcome pre-training} enables our models to learn from unlabelled sources and \textit{ICD+} incorporates hierarchical and textual ICD representations into our models. For future work, we suggest considering pre-trained language models with larger context sizes \cite{longformer,big-bird} and languages other than English \cite{brazilian-notes}. We further encourage work on semantic encoding of negated terms and numerical data from clinical text.

%% file: sections/Appendix.tex
\section{Distribution of Diagnosis and Procedure Labels}
\begin{figure}[t]
  \centering
  \includegraphics[width=82mm]{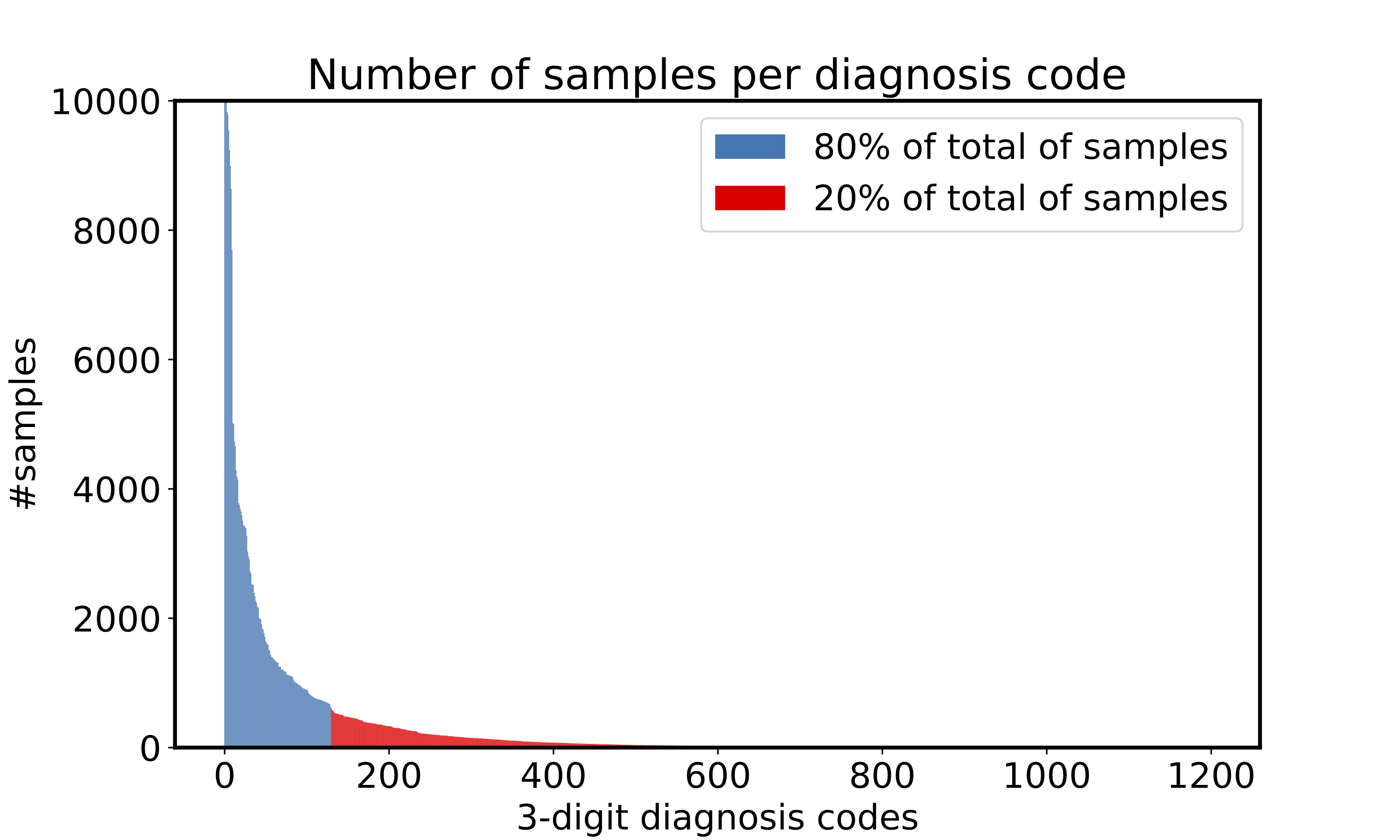}
  \caption{Distribution of ICD-9 diagnosis codes in MIMIC III training set.}
\label{fig:dia_dis}
\end{figure}

\begin{figure}[t]
  \centering
  \includegraphics[width=82mm]{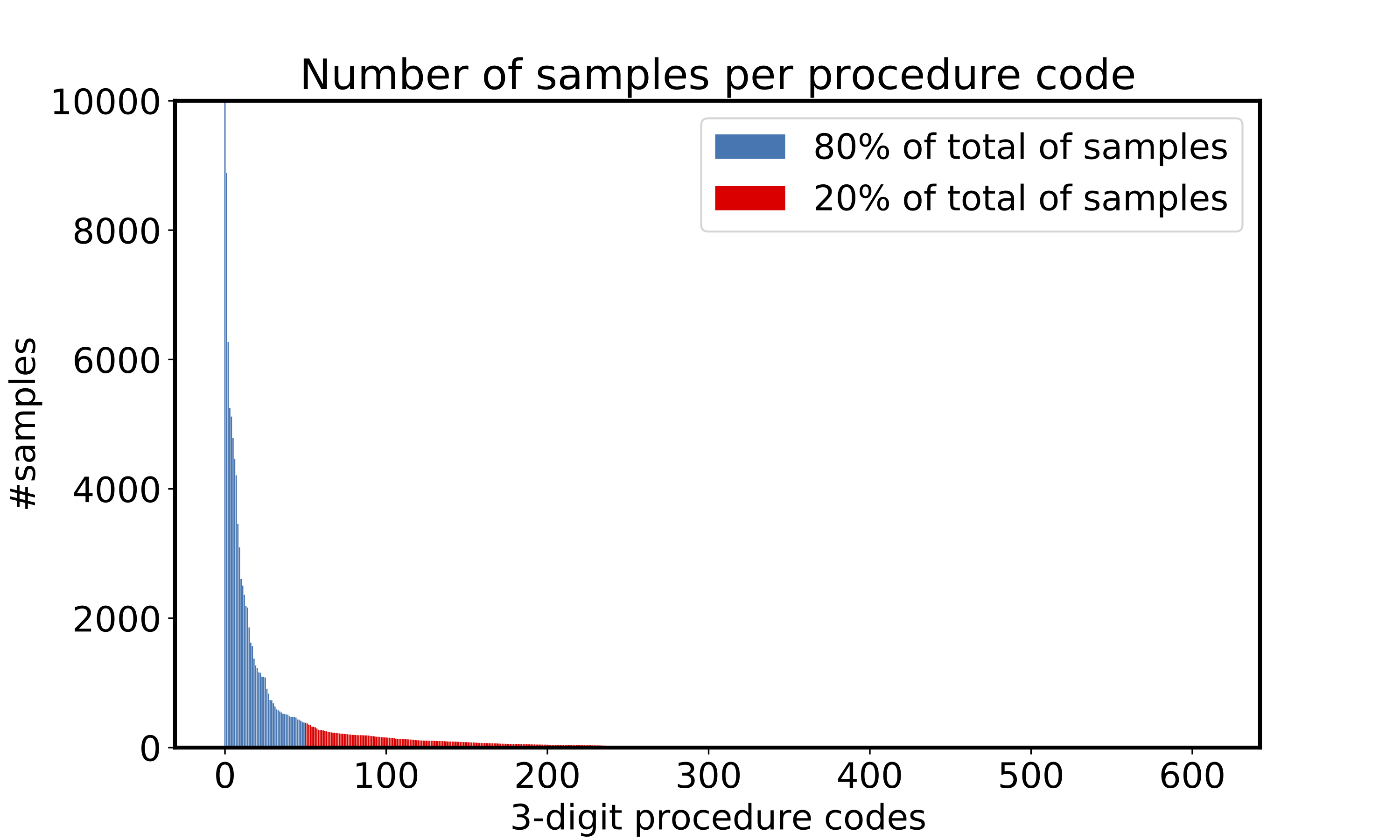}
  \caption{Distribution of ICD-9 procedure codes in MIMIC III training set.}
\label{fig:pro_dis}
\end{figure}

Figure \ref{fig:dia_dis} and Figure \ref{fig:pro_dis} show the distributions of labels in the diagnosis and procedure prediction training sets. Both distributions follow the power law with a long tail of rare codes.

\section{Pre-Processing Clinical Notes}

\subsection{Admission Notes From Discharge Summaries}
\label{app:filter}
We use MIMIC III discharge summaries that contain aggregated information about a patient such as doctor's assessments, relevant lab values, medications, and the patient's history. 
In order to filter the documents by admission sections, we first split all discharge summaries into sections with simple pattern matching. Together with clinical professionals, we then evaluated discharge summaries and identified sections that are known at admission time. We remove all other sections and thus hide information about the further hospital course and discharge of a patient. We exclude notes that do not contain any of the admission sections.
We further apply a patient-wise split into train, validation and test set with a 70/10/20 ratio. 

\subsection{Converting i2b2 Data into Admission Discharge Task}
\label{app:i2b2-prep}
The i2b2 De-identification and Heart Disease Risk Factors Challenge \cite{i2b2-2014,i2b2-2014-2} introduced a dataset that contains clinical notes and discharge summaries annotated based on risk factors and disease indicators. We convert the data into an \textit{admission to discharge} task by selecting five of the annotated conditions which correspond to ICD-9 codes as our labels, namely Hypertension (401), Hyperlipidemia (272), Coronary artery disease (414), Diabetes mellitus (250) and Obesity (278). Just like the MIMIC III diagnosis task, samples are annotated in a multi-label fashion.
In order to convert the clinical notes to admission notes, we use the dataset from \citet{i2b2-section-prediction} that contain section labels per sentence. We then exclude sections that are not known at admission time concurrent to Section 3.2.

\begin{figure}[t!]
  \centering
  \includegraphics[width=78mm]{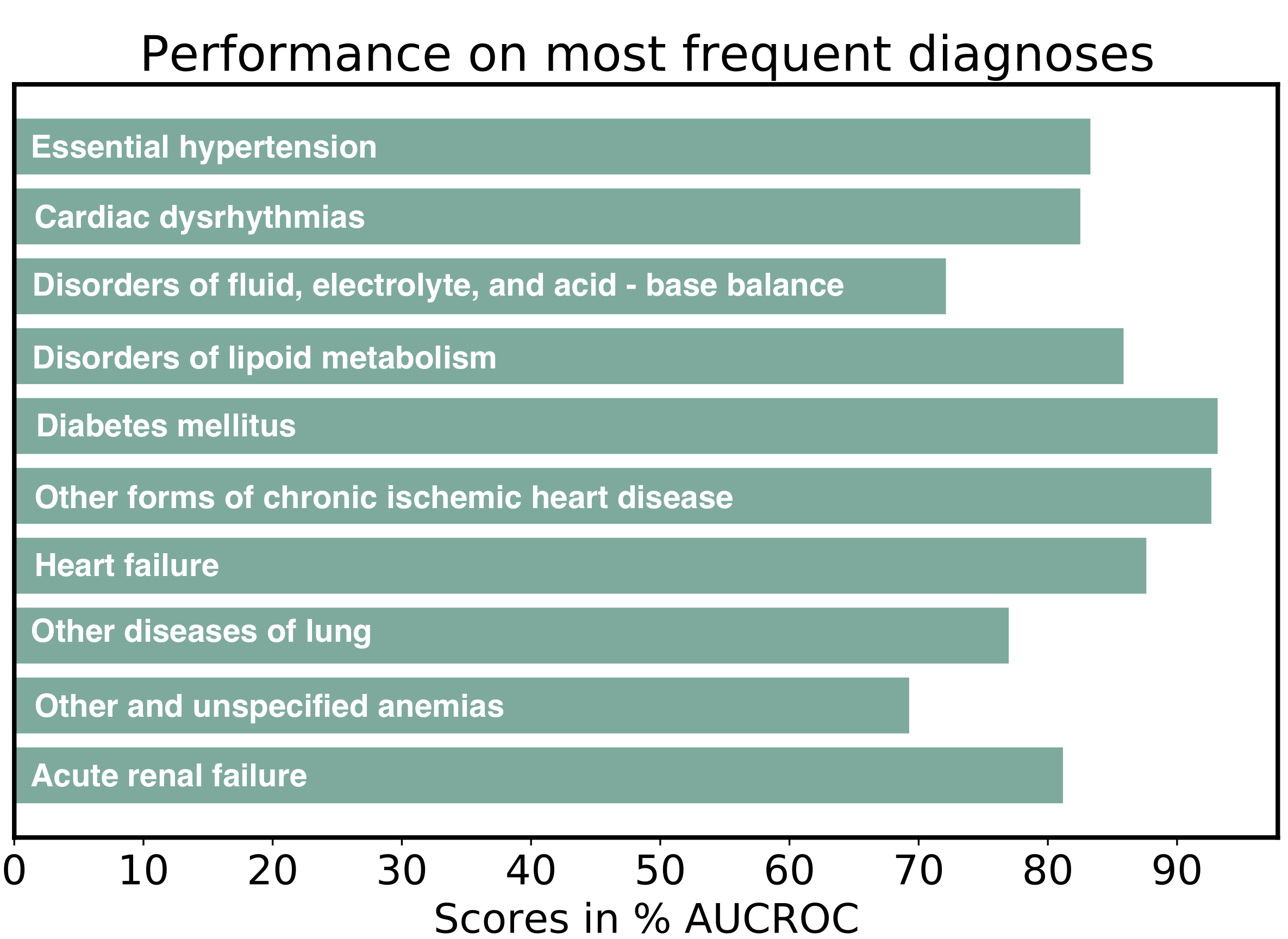}
  \caption{Top 10 diagnoses by frequency with the scores reached by the \textit{CORe All} model.}
\label{fig:top_dia}
\end{figure}

\begin{figure}[t!]
  \centering
  \includegraphics[width=78mm]{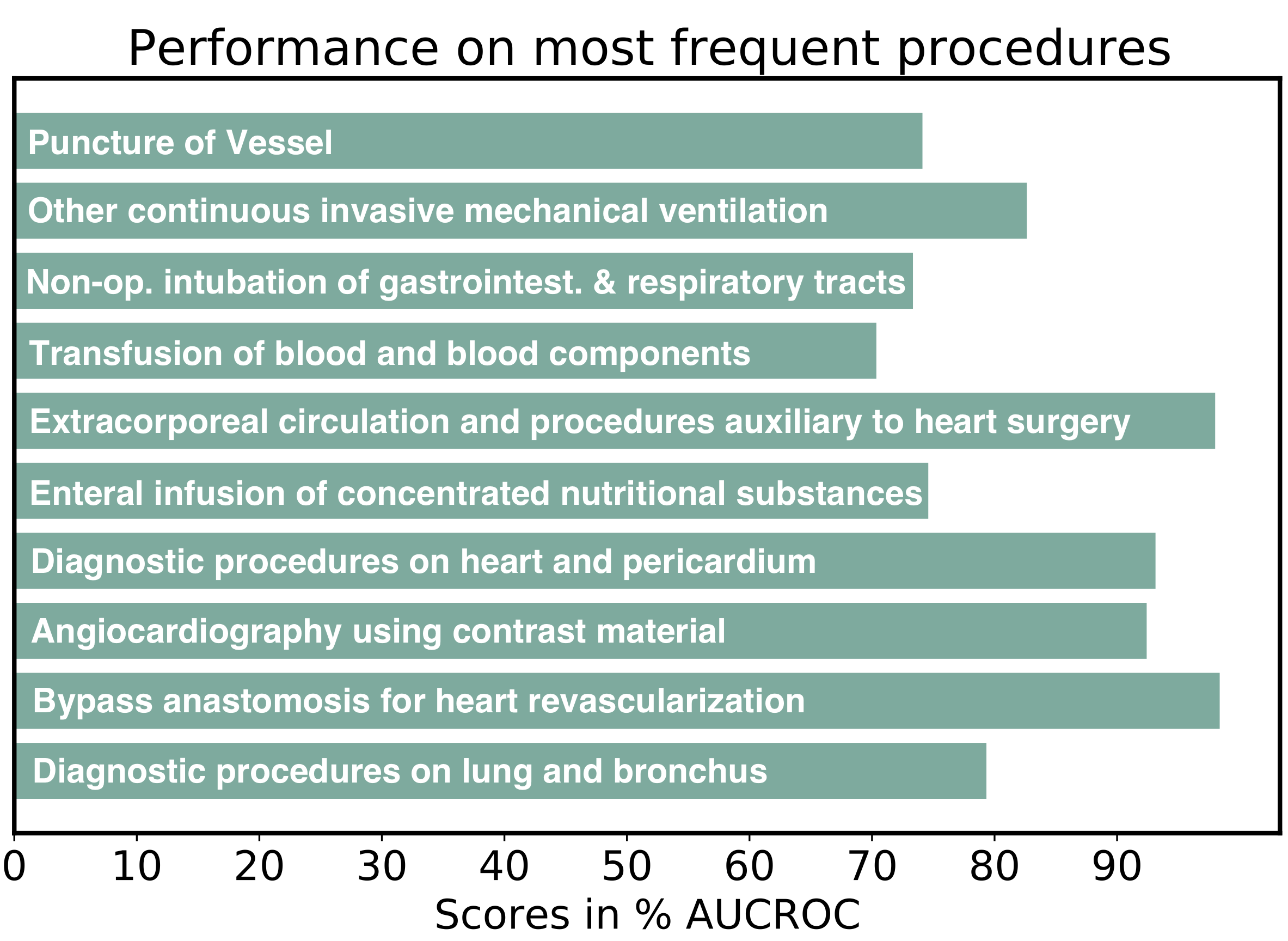}
  \caption{Top 10 procedures by frequency with the scores reached by the \textit{CORe All} model.}
\label{fig:top_pro}
\end{figure}

\section{Hyperparameter Setting}
\label{app:hyperparameter}
We use the following setting for pre-training and fine-tuning of the introduced Transformer-based models:
We use early stopping and apply a random search for tuning the following hyperparameters on the validation set: learning rate [1e-4$-$1e-6], warmup steps [50$-$30k], dropout [0.1$-$0.3], class balancing [True/False] (fine-tuning only), gradient accumulation [1$-$200] with a batch size of 20.

\section{Results on Top 10 Diagnoses and Procedures}
Figures \ref{fig:top_dia} and \ref{fig:top_pro} show the \% AUROC scores of our \textit{CORe All} model on the most frequent labels within the diagnosis and procedure prediction tasks. Figure \ref{fig:top_dia} show that many chronic diseases such as \textit{Essential Hypertension} or \textit{Chronic ischemic heart disease} are among the most common within the MIMIC III dataset and present with relatively high AUROC values. We also observe that very specific codes such as \textit{Diabetes mellitus} and \textit{Bypass Anastomosis} are predicted more easily compared to more general codes such as \textit{Other and unspecified anemias}.\\
Figure \ref{fig:top_pro} further shows the negative influence of inconsistent labeling on standard procedures such as \textit{Puncture of Vessel}. 